\documentclass[conference]{IEEEtran}
\IEEEoverridecommandlockouts
\usepackage{cite}
\usepackage{amsmath,amssymb,amsfonts}
\usepackage{algorithm}
\usepackage{algpseudocode}
\usepackage{subfigure}
\usepackage{textcomp}

\usepackage{graphicx}
\usepackage{xcolor}
\usepackage{enumitem}
\usepackage{amsthm}
\usepackage{multirow}
\theoremstyle{definition}
\newtheorem{definition}{Remark}

\def\BibTeX{{\rm B\kern-.05em{\sc i\kern-.025em b}\kern-.08em
T\kern-.1667em\lower.7ex\hbox{E}\kern-.125emX}}
\begin{document}

\title{Enhancing Interval Type-2 Fuzzy Logic Systems: Learning for Precision and Prediction Intervals   \\

\thanks{This work was supported by MathWorks\textsuperscript{\textregistered} in part by a Research Grant awarded to T. Kumbasar. Any opinions, findings, conclusions, or recommendations expressed in this paper are those of the authors and do not necessarily reflect the views of the MathWorks, Inc.}
}

\author{\IEEEauthorblockN{Ata Köklü} 
\IEEEauthorblockA{\textit{Control and Automation Eng. Dept.} \\
\textit{Istanbul Technical University}\\
Istanbul, Türkiye \\
koklu18@itu.edu.tr}\\
\and
\IEEEauthorblockN{Yusuf Güven} 
\IEEEauthorblockA{\textit{Control and Automation Eng. Dept.} \\
\textit{Istanbul Technical University}\\
Istanbul, Türkiye \\
guveny18@itu.edu.tr}\\
\and
\IEEEauthorblockN{ Tufan Kumbasar} 
\IEEEauthorblockA{\textit{Control and Automation Eng. Dept.} \\
\textit{Istanbul Technical University}\\
Istanbul, Türkiye \\
kumbasart@itu.edu.tr}\\
}

\maketitle

\begin{abstract}
In this paper, we tackle the task of generating Prediction Intervals (PIs) in high-risk scenarios by proposing enhancements for learning Interval Type-2 (IT2) Fuzzy Logic Systems (FLSs) to address their learning challenges. In this context, we first provide extra design flexibility to the Karnik-Mendel (KM) and Nie-Tan (NT) center of sets calculation methods to increase their flexibility for generating PIs. These enhancements increase the flexibility of KM in the defuzzification stage while the NT in the fuzzification stage. To address the large-scale learning challenge, we transform the IT2-FLS's constraint learning problem into an unconstrained form via parameterization tricks, enabling the direct application of deep learning optimizers. To address the curse of dimensionality issue, we expand the High-Dimensional Takagi-Sugeno-Kang (HTSK) method proposed for type-1 FLS to IT2-FLSs, resulting in the HTSK2 approach. Additionally, we introduce a framework to learn the enhanced IT2-FLS with a dual focus, aiming for high precision and PI generation. Through exhaustive statistical results, we reveal that HTSK2 effectively addresses the dimensionality challenge, while the enhanced KM and NT methods improved learning and enhanced uncertainty quantification performances of IT2-FLSs.

\end{abstract}

\begin{IEEEkeywords}
interval type-2 fuzzy sets, design flexibility, curse of dimensionality, uncertainty, accuracy, deep learning 
\end{IEEEkeywords}

\section{Introduction}

Generating Prediction Intervals (PIs) is an especially important task in high-risk applications \cite{pearce,A_review_of_uncertainty_quantification_in_DL}. Interval Type-2 (IT2) Fuzzy Logic Systems (FLSs) are promising model structures owing to their ability to represent uncertainty thanks to the extra degree of freedom provided by the footprint of uncertainty in their Fuzzy Sets (FSs) \cite{mendel_kitap}. Learning IT2-FLSs, compared to Type-1 (T1) FLS, arises with challenging issues as a result of a high number of Learnable Parameters (LPs), the Center of Sets Calculation Method (CSCM), and the constraints to be satisfied, i.e. the set definition of IT2-FSs \cite{Beke_more_than_acc}. Yet, most of the learning methods for IT2-FLSs are merely extensions of T1-FLS ones \cite{Recent_advances_in_neuro_fuzzy_system_survey,T2_fuzzy_broad_learning_system,de2020fuzzy,learning_XAI_Metaheuristic_based_T2_FLS,T2_regression_based_FIS,T2_NN_Cluster} and thus focus on optimizing accuracy, i.e. treating the footprint of uncertainty parameters as extra LPs. They also fall short in addressing the design challenges of IT2-FLSs. 

Integrating Deep Learning (DL) into IT2-FLS is a promising research direction as its T1 fuzzy counterparts have shown successful outcomes while addressing the curse of dimensionality issue \cite{xue2023high,wu_layer_norm,A_state-of-the-art_survey}. Yet, their main focus is solely on precision which contradicts the main claims underscoring the necessity of FLS, namely uncertainty representation. Quite recently, a DL-based IT2-FLS has been proposed that explicitly utilizes the Type Reduced Set (TRS) of IT2-FLSs to learn PI \cite{Beke_more_than_acc}. Yet, a solution to the curse of dimensionality issue is not provided. 

This study proposes enhancements to IT2-FLSs to improve their learning performance in generating PIs alongside accuracy when dealing with high-dimensional data. The enhancements focus on 1) increasing the flexibility of CSCMs, 2) alleviating the impact of the curse of dimensionality and 3) introducing the integration of DL with IT2-FLSs.

We start with increasing the learning flexibility of two well-known CSCM methods, the Karnik-Mendel (KM) and Nie-Tan (NT), aimed to generate High-Quality (HQ) PI (high uncertainty coverage with tight bands). We enhanced the KM by providing an LP for the weighting bounds of the TRS within the output calculation part, resulting in Weighted KM (WKM) while the  NT by defining an LP that weights the Upper MFs (UMFs) and Lower MFs (LMFs) in the fuzzification part, leading to the Weighted NT (WNT). The improvements increase the flexibility of KM in the defuzzification stage and NT in the fuzzification stage. Then, we perform tricks to convert the learning problem of IT2-FLSs to an unconstrained one, facilitating straightforward learning through DL optimizers. To address the curse of dimensionality, we first examine the High-Dimensional Takagi-Sugeno-Kang (HTSK) method proposed for T1-FLSs \cite{wu_layer_norm, HTSK} and extract equivalent MFs representing this method. We show that the HTSK approach is mathematically equivalent to scaling the MFs w.r.t. to the input dimension. Then, we expand the equivalent MF to UMF and LMF to use the HTSK for IT2-FLSs, denoted as HTSK2. 

In this study, we also present a DL-based framework to learn IT2-FLS with a dual focus, namely precision and HQ-PIs. We evaluated the effectiveness and efficiency of the proposed enhancements by presenting exhaustive statistical and comparative results. The results show that HTSK2 is an efficient way to mitigate the curse of dimensionality problem while the deployment of WKM or WNT improves the learning performance and uncertainty quantification capabilities. 

\section{Background on IT2-FLS}
\begin{figure}[h]
        \centering
        \includegraphics[trim=0cm 0cm 0cm 0.67cm, clip,width=0.3\textwidth]
        {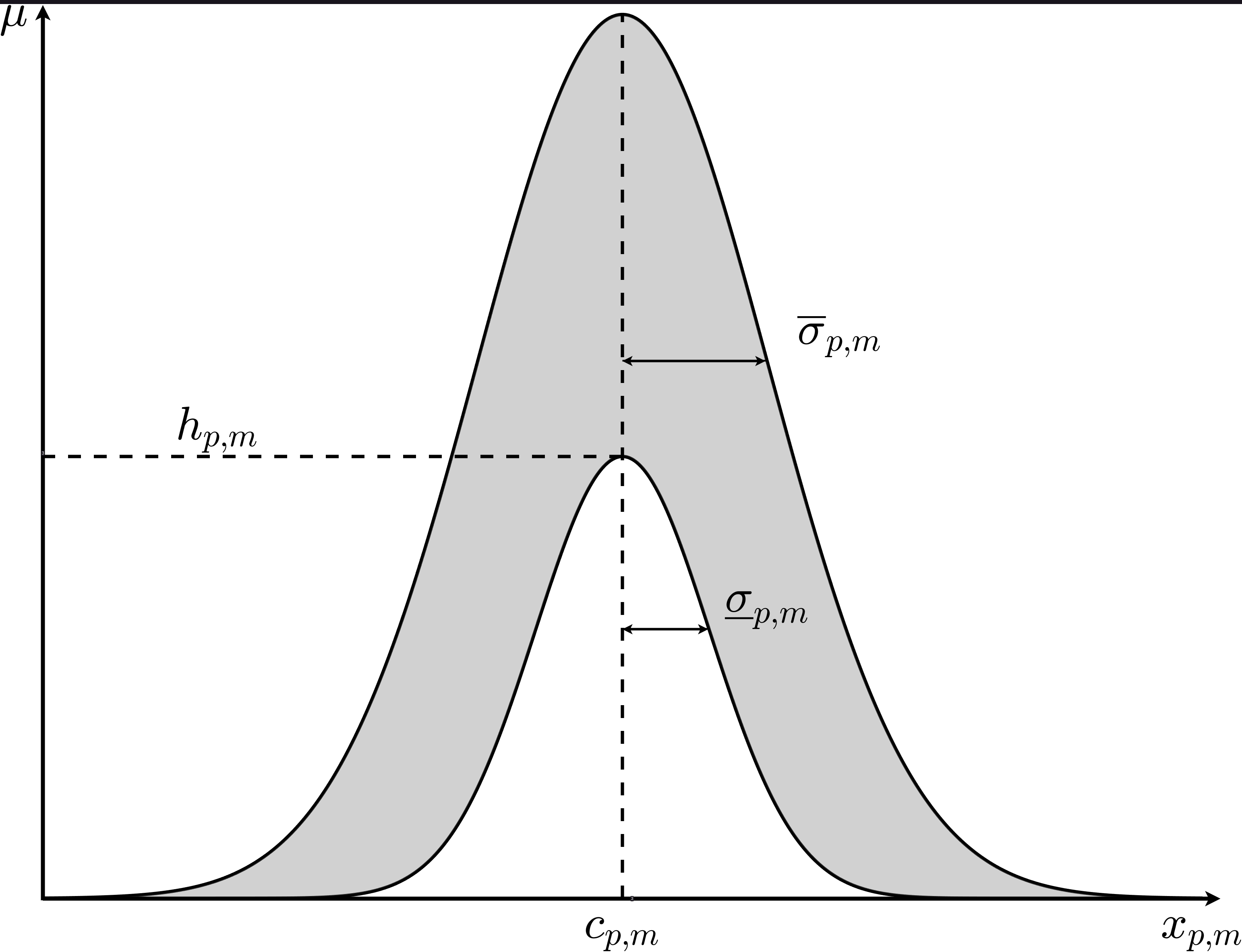}
        \caption{Illustration of an antecedent MF: IT2-FS}
        \label{fig:HS}
    \end{figure}

   The rule structure of an IT2-FLS composed of ${P}$ rules with an input ${\boldsymbol{x} = (x_1,x_2,\ldots,x_M)^T}$ and an output $y$ is as follows:
    \begin{equation}\label{eq:rule}
        R_{p}: \text{If } x_{1} \text{ is } \tilde{A}_{p, 1} \text{ and} \ldots x_{M} \text{ is }\tilde{A}_{p, M} \text{ Then } y \text{ is } \tilde{y}_{p} 
    \end{equation}
    where $y_{p} (p=1,\ldots,P)$ are defined as follows:      \begin{equation}\label{eq:concequent_lower}
    y_{p}=\sum\nolimits_{m=1}^{M} a_{p, m} x_{m}+a_{p, 0}
    \end{equation}
    The antecedent MFs $\tilde{A}_{p, m}$ are described with Gaussian IT2-FSs defined in terms of UMF and LMF as shown in Fig. \ref{fig:HS}. For an input $x_{m}$, the UMF and LMF are defined as:
    \begin{equation}\label{eq:tip2mu}
        \begin{split}
           \overline{\mu}_{\tilde{A}_{p, m}}\left(x_{m}\right)=\exp \left(-\left(x_{m}-\overline{c}_{p, m}\right)^{2} / 2 (\overline{\sigma}_{p, m})^{2}\right)  \\
            \underline{\mu}_{\tilde{A}_{p, m}}\left(x_{m}\right)=h_{p, m} \exp \left(-\left(x_{m}-\underline{c}_{p,m}\right)^{2} / 2 (\underline{\sigma}_{p, m})^{2}\right)
        \end{split}
    \end{equation}
    \noindent where $c_{p, m}$ is the center, $\underline{\sigma}_{p, m}$ and $\overline{\sigma}_{p, m}$ are the standard deviations, while $h_{p, m}$ defines the height of LMF. 

    The output $y \in \tilde{Y}$ is defined via the TRS $\tilde{Y}=[\underline{y}, \overline{y}]$ of the IT2-FLS is obtained through\cite{mendel_kitap}: 
  \begin{equation}\label{eq:tip2 crisp}
        \tilde{Y} =  
        \frac
        {\sum_{p=1}^{P}
        {
        \tilde{F}_{p}y_{p}
        }
        }
        {\sum_{p=1}^{P}
        {
        \tilde{F}_{p}
        }
        }
    \end{equation}
    \noindent where $\tilde{F}_{p}\in[\underline{f}_{p},\overline{f}_{p}]$ is the rule firing that is obtained via:   \begin{equation}\label{eq:tildeF}
        \tilde{F}_{p}= F\left([\underline{\mu}_{\tilde{A}_{p, 1}},\overline{\mu}_{\tilde{A}_{p, 1}}],\ldots, [\underline{\mu}_{\tilde{A}_{p, M}}, \overline{\mu}_{\tilde{A}_{p, M}}] \right)      
    \end{equation}      
    The structure of $F(.)$ depends on the CSCM and t-norm operator $\cap$. Note that there are CSCMs that directly calculate $y$ such as the NT, i.e. without explicitly defining  $\tilde{Y}$\cite{mendel_kitap,Analyzing_Differentiable_Fuzzy_Logic_Operators}.
\section{Enhancements for Learning IT2-FLSs}
    During the learning of IT2-FLSs, we present solutions to the following challenges that arise from the dataset size ($N$) and dimension ($M$), and the deployment of the DL optimizer: 
    \begin{itemize}
        \item \textbf{Flexibility of the CSCM:} CSCMs have different computational complexities and flexibility in calculating $y$ and $\tilde{Y}$. We propose variants of KM and NT for learning high-performance IT2-FLS (See Section III.A).
        \item \textbf{Learning via DL optimizers:} We transform the constraint learning problem of IT2-FLSs into an unconstrained one so that a DL optimizer can be used for learning (See Section III.B).
        \item \textbf{Curse of dimensionality:} To deal with the rule firing problem $f_{p}\approx0$ when $M$ is big, we expand the HTSK proposed for T1-FLS to IT2-FLS (See Section III.C).

    \end{itemize}
    \subsection{Increasing the Flexibility of CSCMs}
    Here, we introduce the proposed WKM and WNT CSCMs. 
    \subsubsection{KM and WKM CSCMs} \label{sec:KM CSCM and its variants for DL}
        The KM is known for its extensive computational as a sorting operation is needed and the TRS is obtained iteratively \cite{mendel_kitap}. It has been shown in \cite{revisiting_KM}:
        \begin{itemize}
            \item To obtain $\underline{y}$, the KM iteratively minimizes \eqref{eq:tip2 crisp} by finding the optimal combination of $\underline{f}$ and $\overline{f}$, namely $\underline{f}^{*}$. For a rule, $\underline{f}^{*}_{p}(x_{p})$ is defined as follows:
            \begin{equation}\label{eq:revisiting_KM_firing_strength_lower}
                \underline{f}^{*}_{p}(x_{p})=\underline{u}_{p}\overline{f}_{p}(x_{p})+(1-\underline{u}_{p})\underline{f}_{p}(x_{p})
            \end{equation}
            where $\underline{u}_{p}\in \{0,1\}$ is the binary optimization variable.
            \item To calculate $\overline{y}$, KM iteratively maximizes \eqref{eq:tip2 crisp} by finding an optimal $\overline{f}^{*}_{p}(x_{p}), \forall p$ which is defined as follows: 
            \begin{equation}\label{eq:revisiting_KM_firing_strength_upper}
                \overline{f}^{*}_{p}(x_{p})=\overline{u}_{p}\overline{f}_{p}(x_{p})+(1-\overline{u}_{p})\underline{f}_{p}(x_{p})
            \end{equation}
            where $\overline{u}_{p} \in \{0,1\}, \forall p$ is the binary optimization variable.
        \end{itemize}
        Then, the TRS can be defined via:
        \begin{equation}\label{eq:ylowup_km}
            \underline{y}(\boldsymbol{x}) =\frac{\sum_{p=1}^{P}{\underline{f}^*_{p}(x_{p})y_{p}}}{\sum_{p=1}^{P}{\underline{f}_{p}}^*(x_{p})} \text{ , }
            \overline{y}(\boldsymbol{x}) =\frac{\sum_{p=1}^{P}{\overline{f}^*_{p}y_{p}(x_{p})}}{\sum_{p=1}^{P}{\overline{f}^*_{p}}(x_{p})}
        \end{equation}
        with 
        \begin{equation}\label{eq:Tip2_f_calc}
           \begin{split} \underline{f}_p(\boldsymbol{x})=\underline{\mu}_{\tilde{A}_{p, 1}}\cap\ldots\cap\underline{\mu}_{\tilde{A}_{p, M}} \\
            \overline{f}_p(\boldsymbol{x})=\overline{\mu}_{\tilde{A}_{p, 1}}\cap\ldots\cap\overline{\mu}_{\tilde{A}_{p, M}}
            \end{split} 
        \end{equation}
        where $\cap$ is the t-norm product operator. The defuzzified output is the average of $\underline{y}(\boldsymbol{x})$ and $\overline{y}(\boldsymbol{x})$ \cite{mendel_kitap}. 
        \begin{equation}\label{eq:crisp_av}
            y(\boldsymbol{x})=\left(\underline{y}(\boldsymbol{x})+\overline{y}(\boldsymbol{x})\right) / 2 
        \end{equation}
        \begin{definition}
        There does exist a direct relationship between switching points ${\{L,R\}}$ and the defined optimization variables $\underline{u}_{p} \in \{0,1\}$ and $\overline{u}_{p} \in \{0,1\}, \forall p$ . Please see \cite{revisiting_KM} for details.              
        \end{definition} 
        In this study, we propose the WKM that has an extra degree of freedom by introducing an LP  $\beta \in [0,1]$ within the defuzzification stage of KM as follows: 
            \begin{equation}\label{eq:crisp_WKM}
                y(\boldsymbol{x}) =  \beta\underline{y}(\boldsymbol{x})+ (1-\beta)\overline{y}(\boldsymbol{x})
            \end{equation}
            Note that, the complexity of WKM is the same as the KM.   

    
    \subsubsection{NT and WNT CSCMs} \label{sec:NT and its variants for DL}
        In the NT CSCM \cite{NT}, $\tilde{A}_{p, m}$ is reduced to T1-FS $A^{*}_{p,m}$ as: \begin{equation}\label{eq:NT_mu}
            \mu_{A^{*}_{p, m}}\left(x_{m}\right) =  (\underline{\mu}_{\tilde{A}_{p, m}}\left(x_{m}\right)+ \overline{\mu}_{\tilde{A}_{p, m}}\left(x_{m}\right))/2
        \end{equation}
        Then, since $\underline{f}^{*}_{p}=\overline{f}^{*}_{p}$, the output is defined as follows:
        \begin{equation}\label{eq:NT_crisp}
            y(\boldsymbol{x}) =  {\frac{\sum\nolimits_{p=1}^{P}  {[\underline{f}_{p}(\boldsymbol{x})y_{p} + \overline{f}_{p}(\boldsymbol{x})y_{p}]} }{\sum\nolimits_{p=1}^{P}  {\underline{f}_{p}(\boldsymbol{x})}  + \sum\nolimits_{p=1}^{P}  {\overline{f}_{p}(\boldsymbol{x})}}}
        \end{equation}
        and an equivalent TRS can be extracted as follows\cite{Beke_more_than_acc}:       \begin{equation}\label{eq:NT_lower}
        \begin{split}
            \underline{y}(\boldsymbol{x}) =  {\frac{2\sum\nolimits_{p=1}^{P}  {\underline{f}_{p}(\boldsymbol{x})\underline{y}_{p}}}{\sum\nolimits_{p=1}^{P}  {\underline{f}_{p}(\boldsymbol{x})}  + \sum\nolimits_{p=1}^{P}  {\overline{f}_{p}(\boldsymbol{x})}}}
            \\
            \overline{y}(\boldsymbol{x}) =  {\frac{2\sum\nolimits_{p=1}^{P}  {\overline{f}_{p}(\boldsymbol{x})\overline{y}_{p}}}   {\sum\nolimits_{p=1}^{P}  {\underline{f}_{p}(\boldsymbol{x})}  + \sum\nolimits_{p=1}^{P}  {\overline{f}_{p}(\boldsymbol{x})}}}
        \end{split}
        \end{equation}

        In this study, we propose the WNT CSCM by introducing an LP $\beta\in[0,1]$ within the fuzzification part, like  \cite{siemensci_alman}, as: 
            \begin{equation}\label{eq:crips_SWNT}
                \mu_{A^{*}_{p, m}}\left(x_{m}\right) = \beta\underline{\mu}_{\tilde{A}_{p, m}}\left(x_{m}\right) + (1-\beta)\overline{\mu}_{\tilde{A}_{p, m}}\left(x_{m}\right)
            \end{equation}
            The corresponding output is then:
            \begin{equation}
                {y}(\boldsymbol{x}) =  
                \frac
                {
                \sum\nolimits_{p=1}^{P}  \left[{\beta\underline{f}_{p}(\boldsymbol{x})\underline{y}_{p}+(1-\beta)\overline{f}_{p}(\boldsymbol{x})\overline{y}_{p}}\right]
                }   
                {
                \sum\nolimits_{p=1}^{P}  {\beta\underline{f}_{p}(\boldsymbol{x})}  
                + 
                \sum\nolimits_{p=1}^{P}  {(1-\beta)\overline{f}_{p}(\boldsymbol{x})}
                }
            \end{equation}
            and an equivalent TRS can be extracted as follows:
            \begin{equation}
            \begin{split}
                \underline{y}(\boldsymbol{x}) =  
                \frac
                {
                2\sum\nolimits_{p=1}^{P}  {\beta\underline{f}_{p}(\boldsymbol{x})\underline{y}_{p}}
                }
                {
                \sum\nolimits_{p=1}^{P}  {\beta\underline{f}_{p}(\boldsymbol{x})}  
                + 
                \sum\nolimits_{p=1}^{P}  {(1-\beta)\overline{f}_{p}(\boldsymbol{x})}
                }
                \\
                \overline{y}(\boldsymbol{x}) =  
                \frac
                {
                2\sum\nolimits_{p=1}^{P}  {(1-\beta)\overline{f}_{p}(\boldsymbol{x})\overline{y}_{p}}
                }   
                {
                \sum\nolimits_{p=1}^{P}  {\beta\underline{f}_{p}(\boldsymbol{x})}  
                + 
                \sum\nolimits_{p=1}^{P}  {(1-\beta)\overline{f}_{p}(\boldsymbol{x})}
                }
            \end{split}
            \end{equation}                       Note that the complexity of WNT is the same as the NT.

    \subsection{Learning IT2-FLS via DL Optimizers}
    Here, we provide all the details on how to learn the enhanced IT2-FLSs via built-in optimizers (e.g., ADAM) and automatic differentiation methods provided within DL frameworks, such as Matlab and PyTorch.
        \subsubsection{Learnable Parameter Sets}\label{sec:Learnable Parameters}
        We define and group the LPs of the IT2-FLSs $(\boldsymbol{\theta})$  regarding whether they are defined within the antecedents $\left(\boldsymbol{\theta}_{\boldsymbol{A}}\right)$, consequents $\left(\boldsymbol{\theta}_{\boldsymbol{C}}\right)$ or CSCM part $\left(\theta_{M}\right)$.
        \begin{itemize}
            \item  $\boldsymbol{\theta}_{\boldsymbol{A}}=\{\boldsymbol{c}, \underline{\boldsymbol{\sigma}}, \overline{\boldsymbol{\sigma}}, \boldsymbol{h}\}$, where 
            $\boldsymbol{c}=\left(c_{1,1}, \ldots, c_{P, M}\right)^{T} \in \mathbb{R}^{P \times M}$,
            $\underline{\boldsymbol{\sigma}}=$ $\left(\underline{\sigma}_{1,1}, \ldots, \underline{\sigma}_{P, M}\right)^{T} \in \mathbb{R}^{P \times M}$,
            $\overline{\boldsymbol{\sigma}}=$ $\left(\overline{\sigma}_{1,1}, \ldots, \overline{\sigma}_{P, M}\right)^{T} \in \mathbb{R}^{P \times M}$,
            $\boldsymbol{h}=\left(h_{1,1}, \ldots, h_{P, M}\right)^{T} \in \mathbb{R}^{P \times M}$. $\boldsymbol{\theta}_{\boldsymbol{A}}$ has in total $4 P M$ LPs.       
            \item $\boldsymbol{\theta}_{\boldsymbol{C}}=\left\{\boldsymbol{a}, \boldsymbol{a}_{0}\right\}$, with $\boldsymbol{a}=\left(a_{1,1}, \ldots, a_{P, M}\right)^{T} \in \mathbb{R}^{P \times M}$ and $\boldsymbol{a_{0}}=\left(a_{1,0}, \ldots, a_{P, 0}\right)^{T} \in \mathbb{R}^{P \times 1}$. $\boldsymbol{\theta}_{\boldsymbol{C}}$ has in total $P(M+1)$ LPs.
            \item $\theta_{M}=\beta$, with $\beta \in \mathbb{R}$.
            
        \end{itemize}
        Thus, we define $\boldsymbol{\theta}$ for an IT2-FLS deploying 
        \begin{itemize}
            \item KM/NT: $\boldsymbol{\theta}=\left\{\boldsymbol{\theta}_{\boldsymbol{A}}, \boldsymbol{\theta}_{\boldsymbol{C}}\right\}$. Thus, \#LP is $5PM+P$. 
            \item WKM/WNT: $\boldsymbol{\theta}=\left\{\boldsymbol{\theta}_{\boldsymbol{A}}, \boldsymbol{\theta}_{\boldsymbol{C}}, \theta_{M}\right\}$. \#LP is $5PM+P+1$.
        \end{itemize}
        
    \subsubsection{Parameterization Tricks for DL Optimizers} \label{sec:trick}
        As stated in \cite{Beke_more_than_acc}, the learning problem of IT2-FLS is defined with constraints $\boldsymbol{\theta} \in \boldsymbol{C}$, arising from the FS definitions. Given that DL optimizers are unconstrained techniques, we present tricks to transform $\boldsymbol{\theta}$ to an unbounded search space. 
        
        For the WKM and WNT, we have the constraint $\beta \in \left[0,1\right]$ due to the definition of \eqref{eq:crisp_WKM} and \eqref{eq:crips_SWNT}. We eliminate them by defining a new LP $\beta^{\prime} \in[-\infty, \infty]$ as follows:
        \begin{equation}
        \beta=sigmoid\left(\beta^{\prime}\right)
        \end{equation}   
        For the remaining LPs, we have employed the parameterization tricks presented in \cite{Beke_more_than_acc}.

    \subsection{Curse of Dimensionality: the HTSK2 Method} \label{sec:Curse of Dimensionality}
        The rule firing problem is mainly caused by the product t-norm operator in \eqref{eq:Tip2_f_calc}. To tackle this issue, we first analyze the HTSK \cite{HTSK} and then present the HTSK2 for IT2-FLS. 
        
        The HTSK approach defines the normalized firing strength $f^{N}_p(\boldsymbol{x})$ calculation via a softmax(.) as follows \cite{HTSK}:     
            \begin{equation}\label{eq:wu_normalized_firing_strength}
                f^{N}_p(\boldsymbol{x}) =\frac{\exp{(Z_{p}^{'})}}{\sum\nolimits_{i=1}^{P}\exp{(Z_{i}^{'})}} = softmax({Z_{p}^{'}})
            \end{equation}           
            with
            \begin{equation}\label{eq:wu_z}
                Z_{p}^{'} = \sum\nolimits_{m=1}^{M} z_{p,m}, \quad 
                z_{p,m} =
                \frac{1}{M}
                \frac
                {
                (x_{m}-c_{p,m})^2
                }
                {
                2(\sigma_{p,m})^{2}
                }
            \end{equation}

    
            To expand the applicability of the HTSK to IT2-FLS, we first extract an equivalent $\mu_{A_{p,m}}(x_{m})$  $(\mu^{eq}_{A_{p,m}}(x_{m}))$. This derivation is used as a basis to define $[\underline{\mu}^{eq}_{\tilde{A}_{p,m}}(x_{m}), \overline{\mu}^{eq}_{\tilde{A}_{p,m}}(x_{m})]$ which are to be processed by $F(.)$. Let us start by reformulating the summation in the exponent of $\exp({\sum\nolimits_{m=1}^{M} z_{p,m}})$ in \eqref{eq:wu_normalized_firing_strength} as a multiplication of $\exp(.)$ :
            \begin{equation}\label{eq:wu_normalized_firing_strength_multipform}
                f^{N}_{p}(\boldsymbol{x}) \equiv 
                \frac
                {
                \prod\nolimits^{M}_{m=1} 
                \exp{\left(z_{p,m}\right)}
                }
                {
                \sum\noindent_{i=1}^{P}
                \left(
                \prod\nolimits^{M}_{m=1}
                \exp{\left(z_{i,m}\right)}
                \right)
                }
            \end{equation}
            Then, we can define the following equivalent $f_p(\boldsymbol{x})$  $(f^{eq}_p(\boldsymbol{x}))$: 
        \begin{equation}\label{eq:wu_f_eq}
                f^{eq}_{p}(\boldsymbol{x})=
                \prod\nolimits^{M}_{m=1}
                \mu^{eq}_{A_{p,m}}(x_{m})
            \end{equation}
            
            \noindent with an equivalent $\mu_{A_{p,m}}(x_{m})$  $(\mu^{eq}_{A_{p,m}}(x_{m}))$ as follows:
            \begin{equation}\label{eq:wu_eq_mu}
                \mu^{eq}_{A_{p,m}}(x_{m})= \exp{\left(-\left(x_{m}-c_{p, m}\right)^{2} /2(\sigma^{eq}_{p, m})^{2}\right)}
            \end{equation}        
            \noindent where $\sigma^{eq}_{p, m} = \sigma_{p,m}\sqrt{M},\forall p,m$. Thus, we conclude that the deployed softmax(.) within the HTSK is scaling $\sigma_{p,m}$ w.r.t. the input dimension $M$ to eliminate $f_{p}\approx0 $ problem $\forall p$. 
            Now, based on the derived $\mu^{eq}_{A_{p,m}}(x_{m})$, we define HTSK2 by defining  $\underline{\mu}^{eq}_{\tilde{A}_{p,m}}(x_{m})$ and $\overline{\mu}^{eq}_{\tilde{A}_{p,m}}(x_{m})$ as follows: 
            \begin{equation}\label{eq:wu_eq_mu_HS_upper_lower}
                \begin{split}
                \overline{\mu}^{eq}_{\tilde{A}_{p,m}}(x_{m})=
                \exp{\left(-\left(x_{m}-c_{p, m}\right)^{2} /2(\overline{\sigma}^{eq}_{p, m})^{2}\right)} \\
                \underline{\mu}^{eq}_{\tilde{A}_{p,m}}(x_{m})=
                h_{p,m}\exp{\left(-\left(x_{m}-c_{p, m}\right)^{2} /2(\underline{\sigma}^{eq}_{p, m})^{2}\right)}
                \end{split}
            \end{equation}
            
            \noindent with $\!\underline{\sigma}^{eq}_{p, m}\!=\!\underline{\sigma}_{p,m}\sqrt{M}$ and  $\overline{\sigma}^{eq}_{p, m}\!=\!\overline{\sigma}_{p,m}\sqrt{M}\!$,  $\forall p,m$.

\section{Dual-Focused IT2-FLS: Accuracy and HQ-PI}
    Here, we introduce a DL framework to learn an IT2-FLS capable not only of yielding accurate predictions but also of generating HQ-PIs, i.e., desired uncertainty coverage and tight band \cite{pearce}. {Algorithm~1} provides the pseudo code for training for a dataset $\left\{\boldsymbol{x}_{n}, y_{n}\right\}_{n=1}^{N}$, where $\boldsymbol{x}_{n}=\left(x_{n, 1}, \ldots, x_{n, M}\right)^{T}$.
    
     To learn an IT2-FLS that has high accuracy and generates HQ-PIs, we define a composite loss, including an accuracy-focused $L_{R}(\cdot)$ and an uncertainty-focused part $\ell(\cdot)$, \cite{Beke_more_than_acc} as:       
        \begin{equation} \label{loss}
            \min _{\boldsymbol{\theta} \in \boldsymbol{C}} L=\frac{1}{N}\sum\nolimits_{n=1}^{N}\left[L_{R}\left(\boldsymbol{x}_{n},y_{n}\right) + \ell\left(\boldsymbol{x}_{n}, y_n, \underline{\tau}, \overline{\tau}\right)\right]
        \end{equation}
       Here, $\boldsymbol{C}$ are the constraints that should be handled as defined in Section III.B.
        \begin{algorithm} 
            \caption{DL-based Dual-Focused IT2-FLS}
            \begin{algorithmic}[1]
            \label{alg:integration}
            \small
            \State \textbf{Input:} $N$ training samples $(x_{n},y_{n})^{N}_{n=1}, \varphi = [\underline{\tau}, \overline{\tau}]$
            \State $P$, number of rules 
            \State $mbs$, mini-batch size
            \State \textbf{Output:} Learnable parameter set ${\theta}$
            \State Initialize ${\theta}$;
            \State Perform tricks ${\theta}$; \algorithmiccomment{Sec. III.B}
            \For{\textbf{each } $mbs$ in $N$}
                \State $[\underline{\mu}^{eq}, \overline{\mu}^{eq}] \leftarrow \text{MF}(x; {{\theta}}_{{A}})$ \algorithmiccomment{Eq. \eqref{eq:wu_eq_mu_HS_upper_lower}} 
                \State $[\underline{f}, \overline{f}] \leftarrow F(\underline{\mu}^{eq}, \overline{\mu}^{eq})$ \algorithmiccomment{Eq. \eqref{eq:tildeF}}
                \State $[\underline{y}, \overline{y}, y] \leftarrow \text{CSCM}(\underline{f}, \overline{f}; \left[\theta_C,\theta_M\right])$ \algorithmiccomment{Sec. III.A}
                
                \vspace{0.05cm}
                \State Compute $L$ \algorithmiccomment{Eq. \eqref{loss1}}
                \State Compute the gradient ${\partial L}/{\partial {\theta}}$
                \State Update ${\theta}$ via Adam optimizer
            \EndFor
            
            \State ${\theta}^* = {\arg \min }(L)$
            \State \textbf{Return} $\theta = {\theta}^{*}$
            
            \end{algorithmic}
        \end{algorithm}
       
       For accuracy purposes, we use the following empiric risk function $L_{R}(.)$:\begin{equation}\label{eq:logcosh}
            L_{R(\epsilon)}= \log \left(\cosh \left(\epsilon_{n}\right)\right)
        \end{equation}
        where $\epsilon_{n}=y_{n}-y\left(\boldsymbol{x}_{n}\right)$. For uncertainty coverage purposes, we construct $\ell(.)$ based on two tilted loss functions as follows: 
        \begin{equation}\label{eq:tilted} 
            \ell\left(\boldsymbol{x}_n, y_n,\underline{\tau}, \overline{\tau}\right) = 
            \underline{\ell}\left(\boldsymbol{x}_n, y_n,\underline{\tau}\right) 
            + 
            \overline{\ell}\left(\boldsymbol{x}_n, y_n,\overline{\tau}\right) 
        \end{equation}
        \noindent with
        \begin{equation}\label{eq:tilted_lower}
         \begin{split}  
            \underline{\ell}= \max(\underline{\tau}(y_{n}-\underline{y}(\boldsymbol{x}_{n})),(\underline{\tau}-1)(y_{n}-\underline{y}(\boldsymbol{x}_{n})))\\           \overline{\ell}=\max(\overline{\tau}(y_{n}-\overline{y}(\boldsymbol{x}_{n})),(\overline{\tau}-1)(y_{n}-\overline{y}(\boldsymbol{x}_{n})))
        \end{split}
        \end{equation}   
        \noindent Here, $\underline{\tau}$ represents a lower quantile while $\overline{\tau}$ an upper quantile to learn an expected amount of uncertainty, i.e. $\varphi=[\underline{\tau}, \overline{\tau}]$. 
        
        To sum up, the final composite loss is defined as follows:
        \begin{align}
         L=\frac{1}{N} \sum_{n=1}^{N}\left[L_{R}\left(\epsilon_{n}\right)+\ \ell\left(x_n, y_n,\underline{\tau}, \overline{\tau}\right)\right] \label{loss1}
        \end{align}
        Thus, for instance, we can aim to learn an IT2-FLS that ensures a PI of $I^{\varphi}=99 \%$ (by setting $\varphi=[0.005,0.995]$) via the TRS $[\underline{y}(\boldsymbol{x}_{n}), \overline{y}(\boldsymbol{x}_{n})]$, while minimizing $\epsilon_{n}$ (i.e accuracy) via the output  $y(\boldsymbol{x}_{n})$ of the IT2-FLS.

\section{Comparative Performance Analysis}
    We present exhaustive comparative results to analyze the learning performances of the proposed DL-based IT2-FLS on the high-dimensional White Wine, Parkinson's Motor UPDRS, and AIDS datasets. The investigation encompasses a dual-fold statistical comparison.
        \begin{enumerate}[label=(\roman*)]
            \item An ablation study to examine how the HTSK2 handles the curse of dimensionality problem. 
            \item A comparative performance analysis to examine the provided flexibility via the proposed WKM and WNT.
        \end{enumerate}

        \subsection{Design of Experiments}
        All datasets are preprocessed by Z-score normalization, and we split $70 \%$ of the data into the training set and $30 \%$ of the data into the test set. For each dataset, we trained IT2-FLSs using identical hyperparameters, including mbs and learning rate, for 100 epochs, except for the PM, where we expanded the training to 1000 epochs. The desired PI coverage was set as $99 \% (\varphi=$ $[0.005,0.995])$. The experiments were conducted within MATLAB\textsuperscript{\textregistered} and repeated with 20 different initial seeds for statistical analysis. 
                
        We formulated IT2-FLSs with 4 distinct CSCM and 2 different parametric IT2-FSs which are HS type IT2-FS and H type IT2-FS. The HS type IT2-FS is parameterized as defined in \eqref{eq:tip2mu}. Conversely, the H type IT2-FS is a version of HS type one with the setting $\underline{\sigma}_{p, m} = \overline{\sigma}_{p, m}, \forall p, m$. Thus, the IT2-FLS-H variant has $PM$ less learnable parameter than the IT2-FLS-HS. We configured each IT2-FLS with both the t-norm product operator (PROD) and the HTSK2 approach. To sum up, we trained 16 different IT2-FLSs defined with  $P=5$ rules.
        
        \subsection{Performance Evaluation} 
        We evaluated the performances via Root Mean Square Error (RMSE), PI Coverage Probability (PICP), and PI Normalized Averaged Width (PINAW) \cite{PINAW}. We anticipate learning an IT2-FLS that yields a low RMSE (high accuracy) and attains PICP close to 99\% with a low PINAW value (HQ-PI). To analyze the curse of dimensionality, we reported the count of "Failed to Train experiments" (\#F2T), denoting instances where the IT2-FLSs yielded NaNs after the training process. Unsuccessful learning was characterized by an IT2-FLS producing a PICP $\leq50$ on the testing dataset, and this was reported as the number of "Failed to learn PI models" (\#FPI).

 \begin{table*}[ht]
    \centering
       \caption{Comparative Testing Performance Analysis Over 20 Experiments}
        \begin{center}
        \begin{tabular}
        {ll|cc|cc|cc}
        \hline
        \multicolumn{2}{c}{Dataset}& \multicolumn{2}{c}{White Wine} & \multicolumn{2}{c}{Parkinson's Motor UPDRS} & \multicolumn{2}{c}{AIDS} \\
        \multicolumn{2}{c}{$M\times N$}  & \multicolumn{2}{c}{$11\times4898$} & \multicolumn{2}{c}{$19\times5875$} & \multicolumn{2}{c}{$23\times2139$} \\
        \hline
        &  & PROD & HTSK2 & PROD & HTSK2 & PROD & HTSK2 \\
        \hline
        \hline
        Metrics& CSCM & \multicolumn{6}{c}{IT2-FLS-H}\\
        \hline
        
        RMSE & KM & 82.40($\pm$2.92)&80.89($\pm$1.94) & 88.15($\pm$9.80) & 61.40($\pm$3.55) &72.30($\pm$2.75) &71.13($\pm$2.43)\\
        & WKM &  82.64($\pm$3.43)&81.47($\pm$2.55) & 86.04($\pm$11.32) & \textbf{61.03($\pm$4.56)} &71.99($\pm$3.66) & \textbf{71.55($\pm$3.72)} \\
        & NT &  80.90($\pm$2.43)&80.07($\pm$2.65) & 80.97($\pm$0.24) & 63.96($\pm$4.30) &--- &70.42($\pm$1.95)\\
        & WNT &  \textbf{79.80($\pm$1.77)}&80.04($\pm$2.54) & 85.87($\pm$5.22) & 63.02($\pm$4.55) &70.56($\pm$2.43) &69.50($\pm$2.71)\\
        \hline
        PICP & KM & 97.09($\pm$1.52)& 97.76($\pm$0.86) & 91.60($\pm$4.31) & 98.72($\pm$0.64) &97.00($\pm$1.24) &97.51($\pm$0.98)\\
        & WKM &  97.11($\pm$1.46) &97.81($\pm$0.90) & 91.49($\pm$4.25) & \textbf{98.94($\pm$0.57)} &97.19($\pm$1.36) & \textbf{97.62($\pm$0.82)} \\
        & NT &  97.32($\pm$0.58) &\textbf{98.29($\pm$0.48)} & 83.83($\pm$3.69) & 94.50($\pm$9.42) &--- &95.09($\pm$1.13)\\
        & WNT &  97.23($\pm$0.92) &98.23($\pm$0.47) & 69.80($\pm$14.70) & 94.82($\pm$9.42) &82.11($\pm$1.61) &96.23($\pm$1.00)\\
        \hline
        PINAW & KM &  74.39($\pm$11.40) &72.57($\pm$7.05) & 165.02($\pm$36.49) & 141.79($\pm$21.77) &185.58($\pm$14.15) &188.12($\pm$15.80)\\
        & WKM &  72.92($\pm$11.62) &71.61($\pm$6.95) & 165.35($\pm$40.35) & \textbf{131.98($\pm$17.69)} & \textbf{179.13($\pm$12.19)} &179.91($\pm$10.01)\\
        & NT &  \textbf{59.91($\pm$5.49)} &63.76($\pm$5.75) & 60.49($\pm$6.77) & 71.99($\pm$10.55) &--- &122.92($\pm$15.58)\\
        & WNT &  60.30($\pm$6.17) &63.89($\pm$5.06) & 50.16($\pm$21.43) & 75.65($\pm$12.39) &62.38($\pm$2.23) &140.80($\pm$11.73)\\
        \hline
        \#F2T$|$\#FPI & KM &  0$|$0 & 0$|$0 & 0$|$0 & 0$|$0 & 0$|$0 & 0$|$0\\
        & WKM &  0$|$0 & 0$|$0 & 0$|$0 & 0$|$0 & 0$|$0 & 0$|$0\\
        & NT &  0$|$1 & 0$|$1 & 0$|$18 & 0$|$0 & 0$|$20 & 0$|$0\\
        & WNT &  0$|$2 & 0$|$0 & 1$|$13 & 0$|$0 & 0$|$14 & 0$|$0\\
        \hline
        \hline
        Metrics& CSCM & \multicolumn{6}{c}{IT2-FLS-HS}\\
        \hline
        
        RMSE & KM & 82.62($\pm$2.46) &80.61($\pm$2.05) &74.17($\pm$10.06) &63.81($\pm$2.99) &71.94($\pm$3.52) &72.40($\pm$3.33)   \\
        & WKM &  82.35($\pm$1.90) &80.98($\pm$1.95) &76.19($\pm$12.14) &\textbf{62.25($\pm$3.60)} &\textbf{69.61($\pm$2.33)} &71.37($\pm$4.66)  \\
        & NT & 81.19($\pm$1.13)  &\textbf{79.52($\pm$1.32)} &--- &66.01($\pm$3.55) &--- &69.96($\pm$2.62)  \\
        & WNT &  80.76($\pm$1.18) &79.70($\pm$1.58) &--- &64.28($\pm$1.42) &--- &69.30($\pm$2.34)  \\
        \hline
        PICP & KM & 97.55($\pm$1.59) &98.20($\pm$0.59) &98.20($\pm$2.88) &99.33($\pm$0.44) &98.64($\pm$0.69) &97.62($\pm$0.92)  \\
        & WKM & 97.83($\pm$1.15)  &\textbf{98.32($\pm$0.64)} &97.86($\pm$3.18) &\textbf{99.09($\pm$0.63)} &\textbf{98.85($\pm$0.73)} &98.01($\pm$1.16)  \\
        & NT &  97.34($\pm$1.16) &98.23($\pm$0.45) &--- &95.64($\pm$2.76) &--- &94.60($\pm$2.39)  \\
        & WNT &  97.81($\pm$0.57) &98.12($\pm$0.42) &--- &95.65($\pm$1.72) &--- &96.55($\pm$0.71)  \\
        \hline
        PINAW & KM & 74.35($\pm$9.12)  &70.36($\pm$7.70) &171.68($\pm$39.58) &149.57($\pm$25.68) &210.76($\pm$19.73) &194.39($\pm$15.69)  \\
        & WKM & 74.02($\pm$9.12)  &70.62($\pm$7.27) &165.46($\pm$38.43) &\textbf{147.81($\pm$24.22)} &200.70($\pm$19.04) &\textbf{188.05($\pm$12.97)}  \\
        & NT & \textbf{59.14($\pm$4.61)}  &62.90($\pm$5.38) &--- &79.30($\pm$3.63) &--- &123.13($\pm$13.44)  \\
        & WNT &  59.81($\pm$2.33) &64.20($\pm$5.27) &--- &80.68($\pm$12.83) &--- &142.86($\pm$12.11)  \\
        \hline
        \#F2T$|$\#FPI & KM & 3$|$0 &0$|$0 &6$|$0 &1$|$0 &0$|$0 &0$|$0  \\
        & WKM & 2$|$0 &0$|$0 &9$|$0 &1$|$0 &0$|$0 &0$|$0  \\
        & NT & 0$|$10 &0$|$2 &2$|$18 &0$|$7 &0$|$20 &0$|$0  \\
        & WNT & 0$|$10 &0$|$1 &1$|$19 &0$|$10 &0$|$20 &0$|$0  \\
        \hline
        \multicolumn{8}{l}{ RMSE and PINAW values are scaled by 100.}\\
        \multicolumn{8}{l}{Bold ones indicate the best results. An IT2-FLS is considered the best if its corresponding PICP falls within the range [97,100]}\\
        \multicolumn{8}{l}{--- denotes an absence of a mean value since $\text{\#F2T}+\text{\#FPI}=20$}\\
        \end{tabular}
    \label{tab:H&HS}
    \end{center}
\end{table*}

\begin{figure*}[hbpt]
        \centering
        \subfigure[IT2-FLS-H]
        {
        \includegraphics[width=0.35\textwidth]
                    {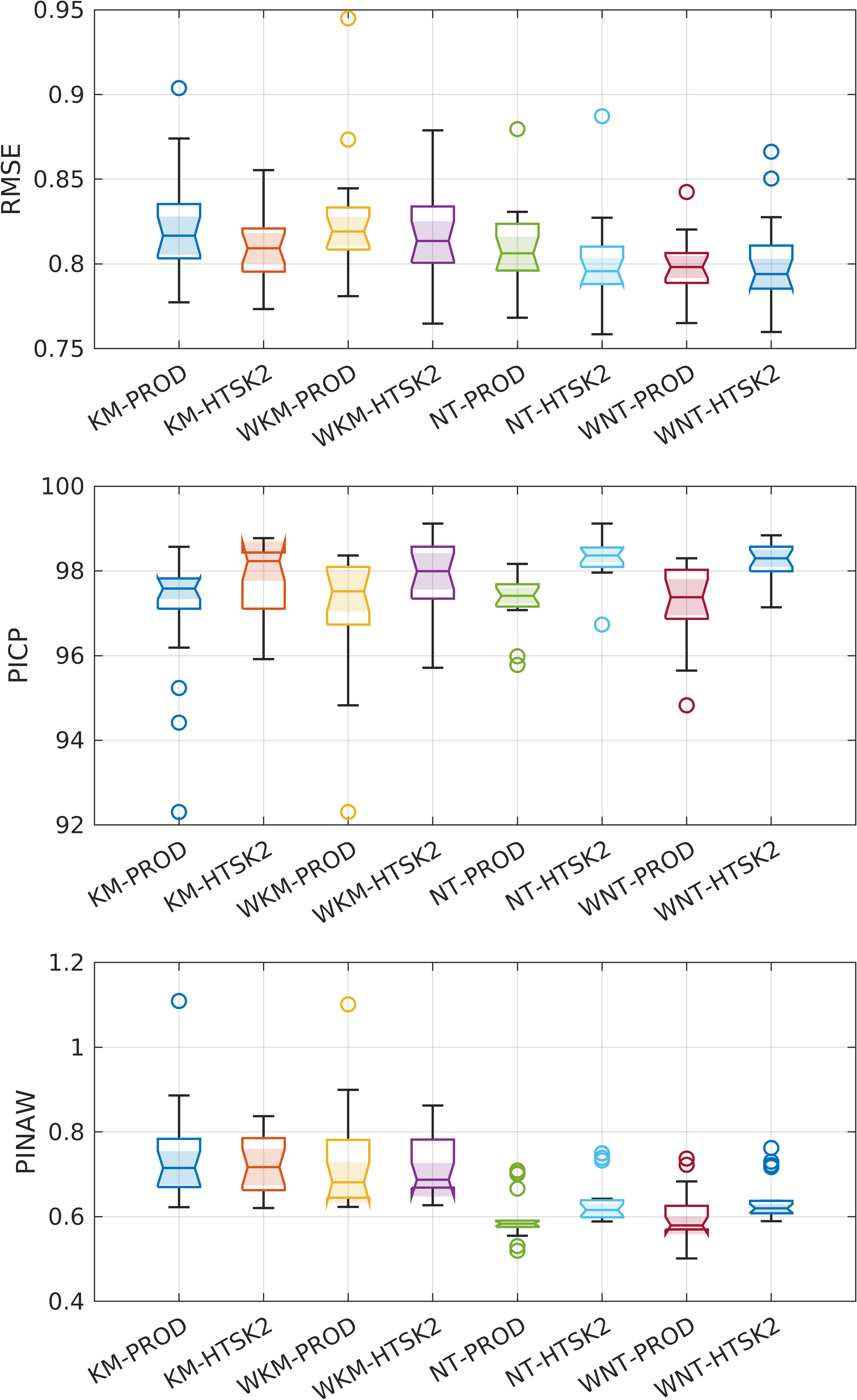}
                    \label{fig:wine_H}
        }
        \subfigure[IT2-FLS-HS]
        {
        \includegraphics[width=0.35\textwidth]
                    {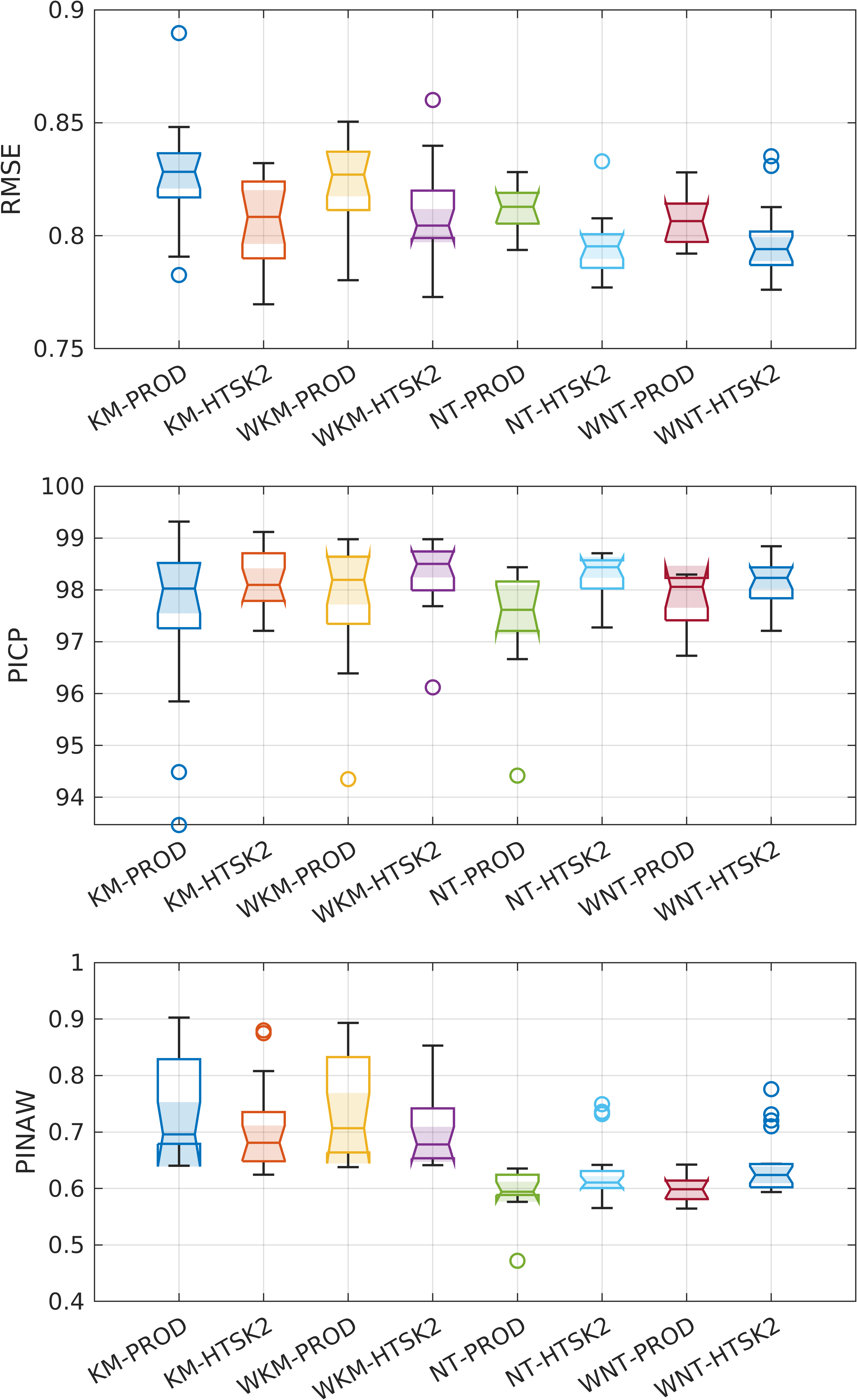}
                    \label{fig:wine_HS}
        
        }
          \caption{White Wine dataset: Notched box-and-whisker plots}
          \label{fig:wine}
\end{figure*}

\begin{figure*}[hbpt]
        \centering
        \subfigure[IT2-FLS-H]
        {
        \includegraphics[width=0.35\textwidth]
                    {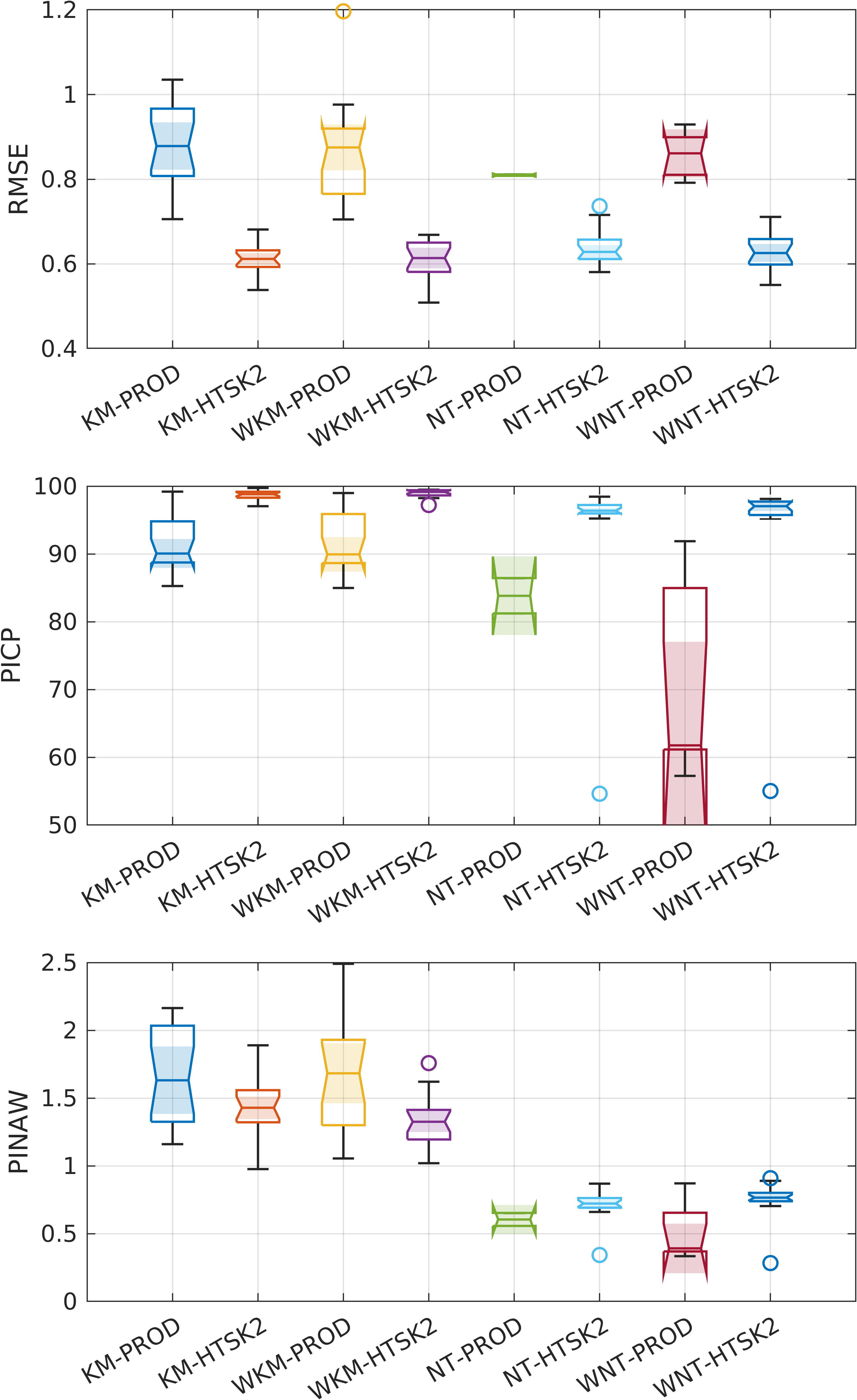}
                    \label{fig:parkinson_H}
        }
        \subfigure[IT2-FLS-HS]
        {
        \includegraphics[width=0.35\textwidth]
                    {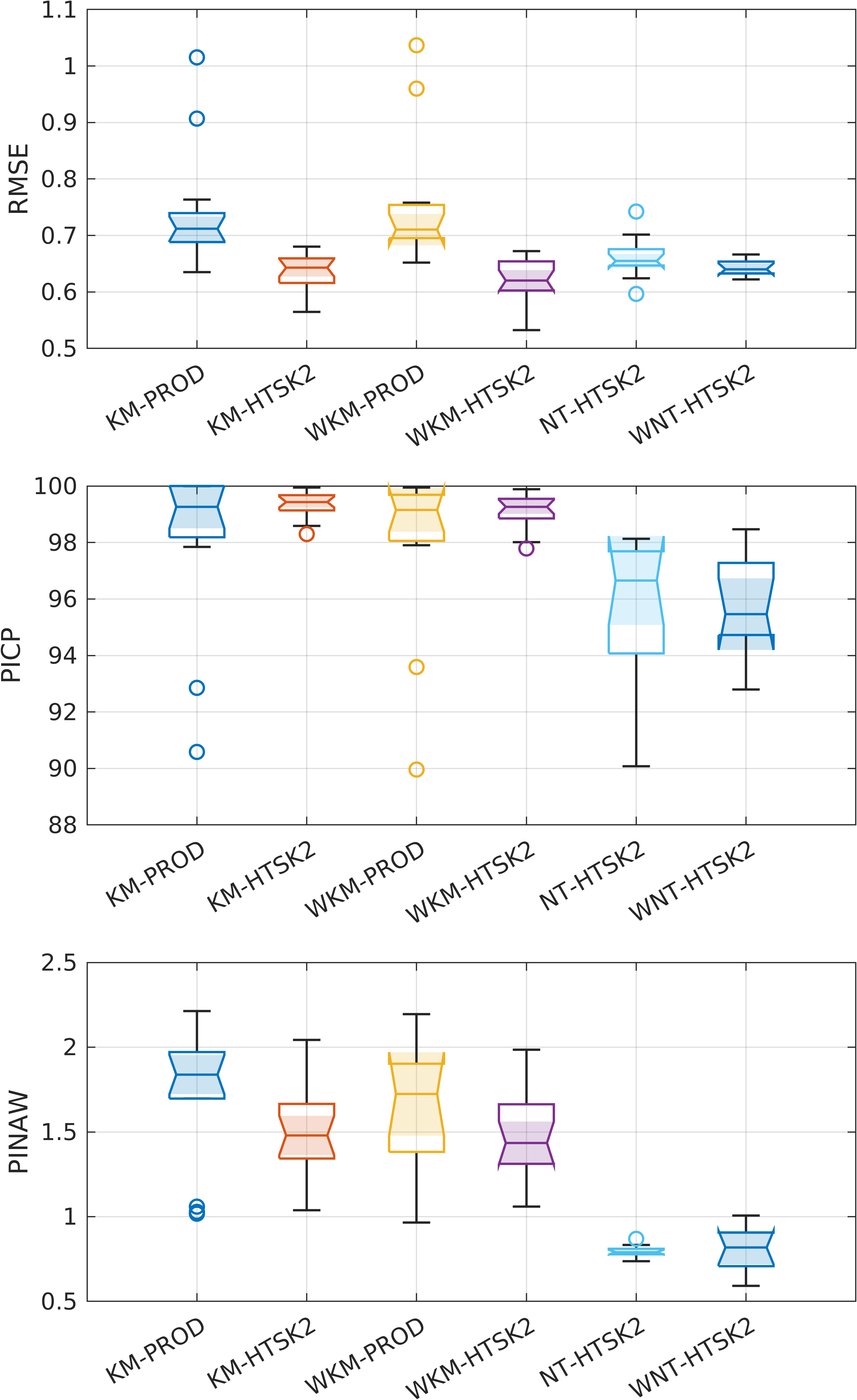}
                    \label{fig:parkinson_HS}
        
        }
          \caption{Parkinson's Motor UPDRS dataset: Notched box-and-whisker plots}
          \label{fig:parkinson}
\end{figure*}

\begin{figure*}[hbpt]
        \centering
        \subfigure[IT2-FLS-H]
        {
        \includegraphics[width=0.34\textwidth]
                    {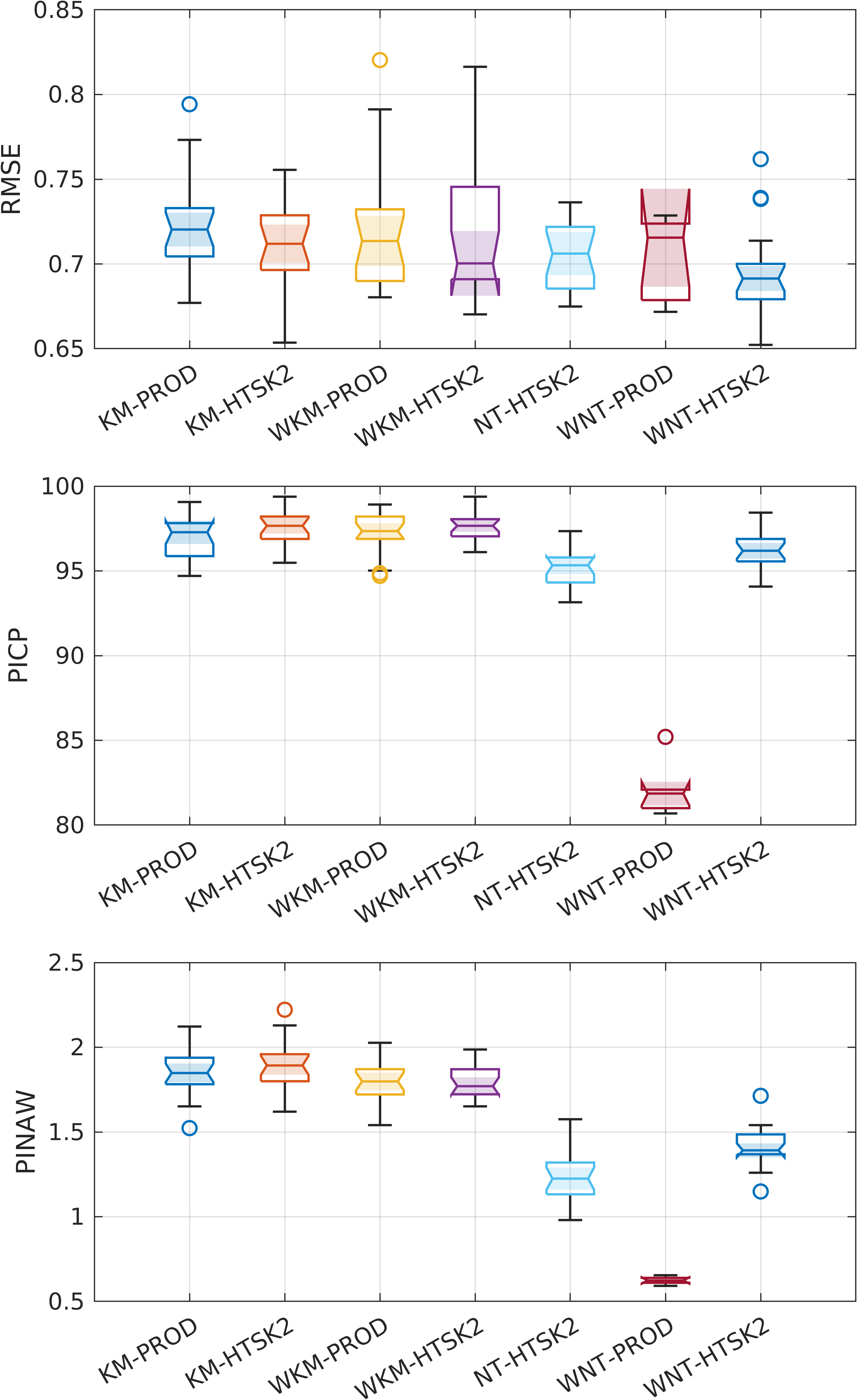}
                    \label{fig:aids_H}
        }
        \subfigure[IT2-FLS-HS]
        {
        \includegraphics[width=0.34\textwidth]
                    {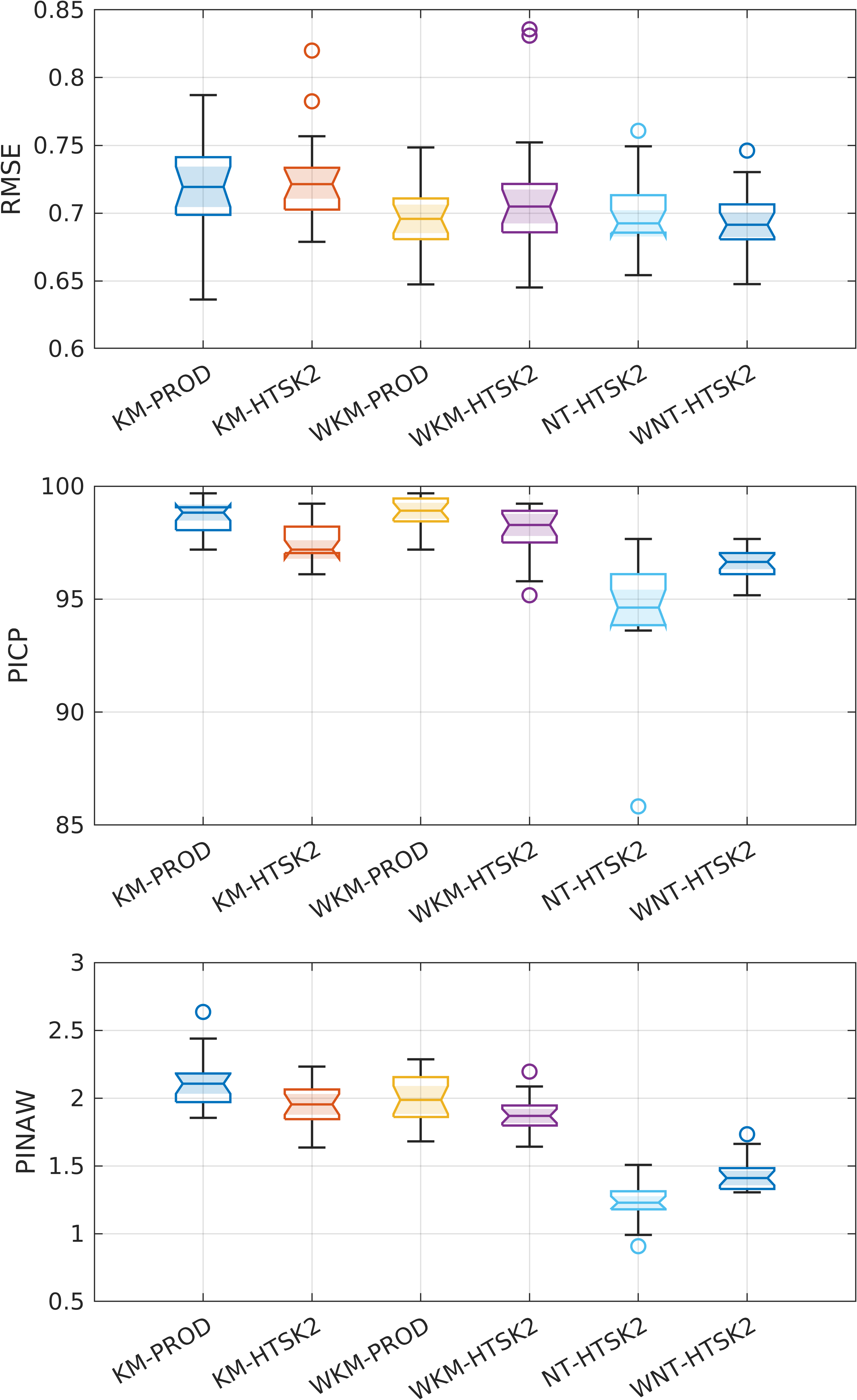}
                    \label{fig:aids_HS}
        
        }
          \caption{AIDS dataset: Notched box-and-whisker plots}
          \label{fig:aids}
\end{figure*}

        Table \ref{tab:H&HS} presents the mean RMSE, PICP, and PINAW and their $\pm1$ standard error of testing data alongside \#F2T and \#FPI  over 20 experiments. In Fig. \ref{fig:wine}---Fig. \ref{fig:aids}, for statistical analysis, provide the notched box and whisker plots obtained from the results of IT2-FLSs showing median (central mark), $25^\text{{th}}$/$75^\text{{th}}$ percentiles (left and right edges of box), i.e. the Inter Quartile Range (IQR), whiskers (line), and outliers (circles). 
        \begin{definition}
        We excluded the results of the IT2-FLSs that F2T and FPI from the performance measures and box plots.
        \end{definition}

        \subsubsection{Curse of Dimensionality}
        Let's examine the performance boost with the deployment of the HTSK2. Observe that:
        \begin{itemize}
            \item The HTSK2 reduced \#F2T experiments as it is particularly evident in the results of the Parkinson dataset.
            \item The HTSK2 significantly decreased \#FPI, resulting in a more robust learning performance. For instance, it can be seen that especially for NT and WNT in Table \ref{tab:H&HS}, the \#FPI is reduced from 20 to 0, yielding nearly 95\% PICP values instead of having an unsuccessful PI. 
            \item For the same \#F2T and \#FPI, the resulting measures of the HTSK2 are better. For instance, as observed from box plots, KM and WKM with HTSK2 resulted in statistically significant performance measures (no intersection of the notches) when compared to their PROD counterparts.
        \end{itemize}
    
    To sum up, IT2-FLSs (regardless of which  CSCM is deployed) defined with the proposed HTSK2 exhibit enhanced capability in addressing the curse of dimensionality. This is evident in the reduction of \#F2T and \#FPI, simultaneous improvement in accuracy, and the generation of HQ-PIs. 
    
    \subsubsection{Flexibility} Let us investigate the performance improvement through the deployment of proposed WKM and WNT CSCMs into IT2-FLSs when utilizing HTSK2. Observe that:      
        \begin{itemize}
            \item WKM has demonstrated superior performance over KM across various datasets and models. As can be observed from the AIDS dataset results, WKM has notably smaller PINAW values and thus has better HQ-PI compared to KM. This trend of superior performance is even more pronounced in the Parkinson dataset, where WKM outperforms KM across all metrics. Also, from the results of both AIDS and Parkinson datasets, WKM has better PICP and PINAW measures compared to KM.
            \item WNT has also demonstrated its superior performance over its NT counterpart. For instance, WNT not only achieves better RMSE and PICP performance measures compared to NT but also exhibits a markedly lower standard deviation in PICP as shown in the results obtained from the AIDS dataset, given in Table \ref{tab:H&HS}.
        \end{itemize}
        
    In the realm of CSCM, both WNT and WKM have excelled in specific metrics. WKM stands out for its superior PICP values, holding the distinction of having the best RMSE in four out of six cases, the best PICP in five out of six cases, and the best PINAW in four out of six cases. This underscores its robustness in providing reliable HQ-PI. Thus, if one were to recommend a CSCM, it would be the WKM, particularly when deployed to the IT2-FLS-HS as it achieves outstanding PICP values, which is crucial in high-risk problems. 

\section{Conclusion \& Future Work}
        This study proposes enhancements for learning IT2-FLSs, focusing on flexibility and mitigating the curse of dimensionality. We proposed two novel CSCMs intending to increase flexibility and achieve HQ-PI. Then, to utilize DL techniques, we alleviated the constraints in IT2-FLS via tricks. We also addressed the curse of dimensionality issue by expanding the HTSK to IT2-FLS. To show the efficiency of the enhancements, we presented comparative analyses and showed that the WKM/WNT and HTSK2 drastically improved the performance of the dual-focused IT2-FLS. The findings underscore the effectiveness of the proposed approaches in enhancing the uncertainty quantification and performance of IT2-FLSs.
        
        As for our future work, we plan to increase the learning capability of IT2-FLSs with further enhancements.
        
\section*{Acknowledgment}
The authors acknowledge using ChatGPT to refine the grammar and enhance the English language expressions.

\bibliographystyle{IEEEtran}
\bibliography{IEEEabrv,ref}
\end{document}